\documentclass{article}

\usepackage{microtype}
\usepackage{graphicx}
\usepackage{subcaption}
\usepackage{booktabs} 

\usepackage{hyperref}

\usepackage{algorithm}
\usepackage{algorithmic}

\usepackage[accepted]{icml2024}

\usepackage{amsmath}
\usepackage{amssymb}
\usepackage{mathtools}
\usepackage{amsthm}

\usepackage{multirow}

\usepackage[capitalize,noabbrev]{cleveref}

\theoremstyle{plain}

\theoremstyle{definition}

\theoremstyle{remark}

\usepackage[textsize=tiny]{todonotes}

\icmltitlerunning{AsymGQA}

\begin{document}

\twocolumn[
\icmltitle{Optimised Grouped-Query Attention Mechanism for Transformers}



\icmlsetsymbol{equal}{*}

\begin{icmlauthorlist}
\icmlauthor{Yuang Chen}{cam}
\icmlauthor{Cheng Zhang}{ic}
\icmlauthor{Xitong Gao}{siat}
\icmlauthor{Robert D. Mullins}{cam}
\icmlauthor{George A. Constantinides}{ic}
\icmlauthor{Yiren Zhao}{ic}
\end{icmlauthorlist}

\icmlaffiliation{cam}{Department of Computer Science, University of Cambridge, Cambridge, United Kingdom}
\icmlaffiliation{ic}{Department of Electrical and Eletronic Engineering, Imperial College London, London, United Kingdom}
\icmlaffiliation{siat}{Shenzhen Institute of Advanced Technology, Chinese Academy of Sciences, Guangdong, China}

\icmlcorrespondingauthor{Yuang Chen}{yc538@cam.ac.uk}
\icmlcorrespondingauthor{Cheng Zhang}{cheng.zhang122@imperial.ac.uk}
\icmlcorrespondingauthor{Xitong Gao}{xt.gao@siat.ac.cn}
\icmlcorrespondingauthor{Robert D. Mullins}{robert.mullins@cl.cam.ac.uk}
\icmlcorrespondingauthor{George A. Constantinides}{g.constantinides@imperial.ac.uk}
\icmlcorrespondingauthor{Yiren Zhao}{a.zhao@imperial.ac.uk}

\icmlkeywords{Machine Learning, ICML}

\vskip 0.3in
]



\printAffiliationsAndNotice{\icmlEqualContribution} 

\begin{abstract}

Grouped-query attention (GQA) has been widely adopted in LLMs to mitigate the complexity of multi-head attention (MHA). 
To transform an MHA to a GQA, neighbour queries in MHA are evenly split into groups where each group shares the value and key layers. 
In this work, we propose AsymGQA, an activation-informed approach to asymmetrically grouping an MHA to a GQA for better model performance.
Our AsymGQA outperforms the GQA within the same model size budget. For example, AsymGQA LLaMA-2-7B has an accuracy increase of 7.5\% on MMLU compared to neighbour grouping. Our approach addresses the GQA’s trade-off problem between model performance and hardware efficiency.
\end{abstract}

\section{Introduction}
\label{submission}

\begin{figure}[t]
\centering
\includegraphics[width=\linewidth]{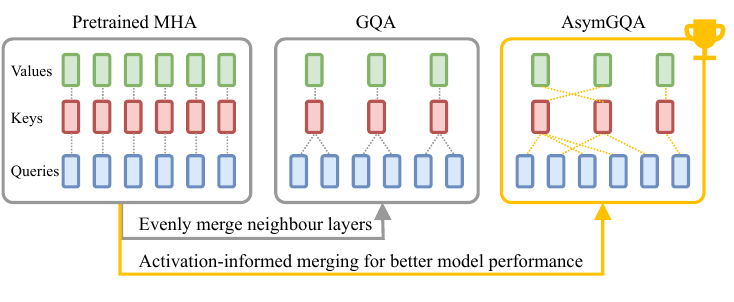}
\vspace{-2em}
\caption{Comparison of GQA and AsymGQA. AsymGQA leverages activation-induced layer similarity to determine the attention head grouping for better model performance.}
\label{fig:motivation}
\end{figure}

Transformer-based models have achieved remarkable success on large-scale language tasks~\cite{devlin2019bert,vaswani2023attention,brown2020language}. Multi-head attention (MHA), the core operation of the Transformer, allows the model to attend to information from different representation subspaces at different positions. However, computational and memory complexity increases quadratically with the sequence length in MHA. To mitigate this problem, researchers have introduced grouped-query attention (GQA)~\cite{ainslie2023gqa}, which evenly splits query heads into groups, and each group shares a single key and value layer (see GQA in \Cref{fig:motivation}).

GQA trades model quality for hardware efficiency. By operating on groups rather than individual queries, GQA reduces the computation cost in Transformer and consumes less memory. 
Typically, GQA models are created by design -- they are originally designed and trained explicitly as GQA models. In this study, we \textbf{\textit{investigate the challenges of converting MHA into a GQA}}, essentially treating GQA as an post-training optimization for the efficient deployment of LLMs.
The naive merging is then take the average of all key and value layers in a group.
Such a simple merging views every head in MHA equally, but our experiments show that this can cause a significant quality degradation even after fine-tuning. To mitigate the quality degradation, we propose asymmetric GQA (AsymGQA), an activation-informed merging approach that considers similarity between layers. Specifically, our contributions are as follows:

\begin{itemize}
    \item We introduce AsymGQA, an activation-informed fusion approach for converting MHA into GQA models, which delivers superior model performance within the same computational constraints.
    \item Through extensive experiments, we answer two unexplored GQA questions. We first verify that our activation-induced method improves the performance for evenly grouped GQA models. Furthermore, we find that more performance gain could be achieved if asymmetric grouping (varying group size) is allowed.     
\end{itemize}

AsymGQA models significantly outperform the GQA baseline. For example, LLaMA-2-7B with an average group size of 4 has an increase in accuracy of 7.2\% on MMLU compared to naive MHA to GQA conversion.



\section{Method}

\Cref{sec:method:grouping-strategies} introduces our search-based grouping methods. \Cref{sec:method:similarity} elaborates how we calculate the similarity of key (value) layers to guide the search.

\begin{figure}[t]
     \centering
     \begin{subfigure}[b]{\linewidth}
         \centering
         \includegraphics[width=\textwidth]{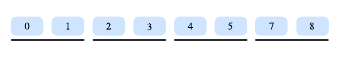}
         \caption{Neighbour grouping (baseline).}
         \label{fig: nb}
     \end{subfigure}
     \hfill
     \begin{subfigure}[b]{\linewidth}
         \centering
         \includegraphics[width=\textwidth]{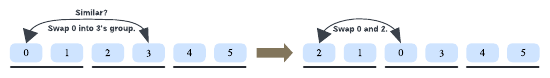}
         \caption{Activation-informed symmetric grouping.}
         \label{fig: sg}
     \end{subfigure}
     \hfill
     \begin{subfigure}[b]{\linewidth}
         \centering
         \includegraphics[width=\textwidth]{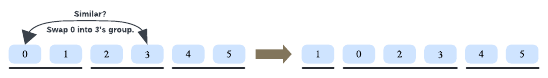}
         \caption{Activation-informed asymmetric grouping.}
         \label{fig: ag}
     \end{subfigure}
        \caption{Naive neighbour grouping vs AsymGQA.}
        \label{fig:three graphs}
\end{figure}


\subsection{Grouping Strategies}

\label{sec:method:grouping-strategies}
\paragraph{Neighbour grouping} We propose a naive scheme called neighbour grouping. Neighbour grouping clusters adjacent attention heads (key layers and value layers) together, while keeping all the groups equally-sized. \Cref{fig: nb} provides a visualization of neighbour grouping. This serves as the baseline in this work. 

\paragraph{Activation-informed symmetric grouping}

We apply the grouping sequentially from the initial MHA layer to the final one. In each MHA layer, the key and value layers are grouped independently. Our proposed method employs a search strategy to determine the optimal grouping of key (and value) layers based on the similarity among them within MHA, as detailed in Section~\ref{sec:method:similarity}. 

As illustrated in \Cref{fig: sg}, before the search starts, a key layer grouping $g$ of size $m$ is generated randomly. In each search iteration, $g$ may be updated so that similar heads are swapped into the same group. Specifically, in one iteration, a key layer $a$ is randomly sampled ($a$'s group is noted as $A$), then the similarity between $a$ and all the other key layers is calculated. Among the top-$k$ most similar layers not belonging to group $A$, a key layer $b$ is selected ($b$'s group noted as $B$). Another head $b'$ in group $B$ is sampled to exchange with $a$, to maintain the group size. If the swapped model has a higher accuracy, the best accuracy and grouping are updated. 
We also introduce random noise to the search to explore more grouping possibilities. An acceptance probability is set to update $g$ even if swapped model has a lower accuracy. Furthermore, another probability is specified that resets $g$ at the start of each iteration to prevent the search from being trapped in a local minima. The detailed algorithm is defined in~\Cref{alg:sym-group} in~\Cref{appendix:alg-grouping-search}.

\paragraph{Activation-informed asymmetric grouping (AsymGQA)} Compared to symmetric grouping, {asymmetric grouping} allows for varied group sizes. Asymmetric grouping is particularly promising in scenarios where the relevance of information is not uniformly distributed across the input space. Asymmetric grouping also extends the search space from only containing equal-sized groups to any groupings, providing opportunities to find even better grouping configuration than symmetric grouping (\Cref{fig: ag}).

Support for asymmetric grouping requires only minor changes to the search algorithm. In symmetric grouping, to maintain the idenitical group sizes, whenever an key or value layer is moved to a new group, an element in the new group must be swapped back. To allow for groups with varying sizes, another parameter $p_{preserve}$ is introduced, representing the probability of preserving the group sizes after swap. A larger $p_{preserve}$ encourages a more unbalanced grouping. A detailed description is included in~\Cref{alg:asym-group} in the Appendix \Cref{appendix:alg-grouping-search}.

\subsection{Activation-Informed Head Similarity}
\label{sec:method:similarity}
As stated in~\Cref{sec:method:grouping-strategies}, our search calculates the similarity between key (value) layers at the beginning of each iteration to guide the search. We \textit{cluster similar layers into the same group}, which is advantageous for two reasons:
\begin{itemize}
    \item It leads to better optimization and generalization. As similar layers contribute to a focused gradient update that is consistent across the group, the error signal propagated back through the network can more effectively tune the shared parameters, enhancing the stability and efficiency of learning.
    \item When attention heads share similar and value layers, they are likely to encode and focus on comparable aspects of the input data. This similarity in processing enables more coherent feature extraction, as these heads reinforce each other's understanding and interpretation of specific data patterns, leading to a more nuanced and detailed representation within that particular subspace of the feature space.
\end{itemize}

We measure the similarity between pairs of key (or value) layers using two potential approaches:

\begin{enumerate}
    \item Define the similarity using the difference between the weights of two layers.
    \item Define the similarity using the difference between the output activations of two layers.
\end{enumerate}

Based on our experiments (See~\Cref{appendix:w-vs-act-sim}), we find that activation-informed similarity is a better reference to guide the search. We use consine similarity between vectors to define activation-informed similarity between two layers.

\begin{table*}[ht]
\caption{Comparison between AsymGQA and GQA of LLaMA-2-7B. The column ``GQA'' means the standard GQA which uses neighbour grouping to merge heads. Results of full fine-tuning (``Full FT'') and parameter-efficient LoRA fine-tuning (``LoRA'') after grouping are both included. The highest accuracy is highlighted in \textbf{bold}. Ech cell has two values, Acc ($\Delta$), wehre Acc is accuracy, and $\Delta$ denotes the accuracy difference from fine-tuned MHA, \textit{i.e.}, group size equals 1. We observe that AsymGQA consistently achieves higher accuracy than GQA by a clear margin and close to fine-tuned MHA.}
\vspace{1em}
\centering
\begin{small}
\begin{tabular}{@{}lccccc@{}}
\toprule
\multirow{2}{*}{Task} & \multirow{2}{*}{Group size} & \multicolumn{2}{c}{Full FT}                 & \multicolumn{2}{c}{LoRA}                    \\ \cmidrule(l){3-4}  \cmidrule(l){5-6}
                      &                             & GQA              & AsymGQA                  & GQA              & AsymGQA                  \\ \midrule
\multirow{4}{*}{SST2} & 2                           & 90.9±0.5 (-2.4)  & \textbf{92.9±0.2 (-0.4)} & 90.6±0.5 (-2.6)  & \textbf{92.6±0.1 (-0.6)} \\
                      & 3                           & 88.4±0.3 (-4.9)  & \textbf{92.0±0.4 (-1.3)} & 88.0±0.6 (-5.2)  & \textbf{91.8±0.3 (-1.4)} \\
                      & 4                           & 87.3±0.6 (-5.8)  & \textbf{90.4±0.2 (-2.9)} & 87.1±0.3 (-6.1)  & \textbf{90.1±0.1 (-3.1)} \\
                      & 6                           & 86.8±0.6 (-6.5)  & \textbf{89.6±0.5 (-3.7)} & 86.2±0.3 (-7.0)  & \textbf{89.5±0.1 (-3.7)} \\ \midrule
\multirow{4}{*}{QNLI} & 2                           & 83.7±0.2 (-6.4)  & \textbf{89.5±0.6 (-0.6)} & 82.9±0.2 (-6.7)  & \textbf{89.0±0.3 (-0.6)} \\
                      & 3                           & 80.5±0.6 (-9.6)  & \textbf{88.6±0.4 (-1.5)} & 80.3±0.5 (-9.3)  & \textbf{88.3±0.5 (-1.3)} \\
                      & 4                           & 74.6±0.2 (-15.5) & \textbf{86.5±0.7 (-3.6)} & 73.2±0.2 (-16.3) & \textbf{84.9±0.7 (-3.7)} \\
                      & 6                           & 73.2±0.6 (-16.9) & \textbf{85.0±0.3 (-5.1)} & 72.0±0.2 (-17.6) & \textbf{84.5±0.6 (-5.1)} \\ \midrule
\multirow{4}{*}{MNLI} & 2                           & 79.7±0.4 (-1.5)  & \textbf{81.0±0.1 (-0.2)} & 78.7±0.6 (-2.7)  & \textbf{80.8±0.6 (-0.6)} \\
                      & 3                           & 77.7±0.6 (-3.5)  & \textbf{80.3±0.6 (-0.9)} & 77.0±0.6 (-4.4)  & \textbf{80.2±0.2 (-1.2)} \\
                      & 4                           & 75.1±0.5 (-6.2)  & \textbf{78.4±0.5 (-2.8)} & 74.4±0.3 (-7.0)  & \textbf{78.0±0.4 (-3.3)} \\
                      & 6                           & 72.7±0.3 (-8.6)  & \textbf{77.3±0.1 (-3.9)} & 72.0±0.4 (-9.3)  & \textbf{77.2±0.2 (-4.1)} \\ \midrule
\multirow{4}{*}{MMLU} & 2                           & 34.5±0.5 (-5.3)  & \textbf{39.3±0.7 (-0.5)} & 33.7±0.3 (-6.7)  & \textbf{39.6±0.5 (-0.8)} \\
                      & 3                           & 33.0±0.1 (-6.7)  & \textbf{38.3±0.7 (-1.5)} & 32.7±0.2 (-7.7)  & \textbf{39.0±0.3 (-1.4)} \\
                      & 4                           & 29.0±0.4 (-10.8) & \textbf{36.5±0.2 (-3.3)} & 28.5±0.6 (-11.9) & \textbf{36.8±0.3 (-3.6)} \\
                      & 6                           & 26.4±0.2 (-13.4) & \textbf{33.8±0.2 (-6.0)} & 26.3±0.1 (-14.1) & \textbf{34.7±0.3 (-5.7)} \\ \bottomrule
\end{tabular}
\end{small}
\label{tab: 1}
\end{table*}

The similarity $\mathrm{sim}(A,B)$ between two activation matrices $A\in\mathbb{R}^{n\times m}$ and $B\in\mathbb{R}^{n\times m}$ using row vectors is as follows.

\begin{equation}\label{eq:sim-A-B}
\begin{split} 
    \mathrm{sim}(A,B)=\frac{1}{2}(\sum_{i=0}^{n-1}\max_{j=0}^{n-1} \mathrm{cosim}(A_{i, *}, B_{j, *}) \\+ \sum_{i=0}^{n-1}\max_{j=0}^{n-1} \mathrm{cosim}(B_{i, *}, A_{j, *}))
\end{split}
\end{equation} 

where $\mathrm{cosim}(\cdot)$ calculates the cosine similarity between two vectors $\mathbf{u}$ and $\mathbf{v}$:

\begin{equation}
    \mathrm{cosim}(\mathbf{u}, \mathbf{v}) = \frac{\mathbf{u}\cdot \mathbf{v}}{ \Vert \mathbf{u}\Vert \Vert\mathbf{v}\Vert }
\end{equation}

\section{Evaluation}

\Cref{sec:evaluation:exp-setup} introduces our basic experiment setup. \Cref{sec:evaluation:asymgqa} presents our main results on activation-informed asymmetric grouping, which is the best grouping strategy out of the three introduced in \Cref{sec:method:grouping-strategies}. \Cref{sec:evaluation:ablation-study} includes two ablation studies to verify the efficacy of activation-informed grouping and varied group sizes.

\subsection{Experiment Setup}
\label{sec:evaluation:exp-setup}
\paragraph{Models and datasets} We apply our methods to popular decoder-only models including OPT \cite{zhang2022opt}, LLaMA \cite{touvron2023llama} and LLaMA-2 \cite{touvron2023llama2} with the number of parameters ranging from 125 million to 7 billion. We evaluated these models on QNLI~\cite{wang-etal-2018-glue}, MNLI~\cite{williams2017broad}, SST2\cite{socher2013recursive}, and MMLU \cite{DBLP:journals/corr/abs-2009-03300}. The entire MMLU is evaluated under a zero-shot setting. Each experiment has three independent runs with different random seeds. The mean and standard deviation of three runs are calculated. 

\paragraph{Grouping} For each layer, 10 search iterations are executed with different groupings, with an acceptance probability of $0.1$, a reset probability of $0.1$ and a preservation probability of $0.2$. In each iteration, we randomly choose one of the top 3 closest heads to group.

\paragraph{Fine-tuning} We fine-tune the grouped model for three more epochs to recover the model performance. We include both full fine-tuning and LoRA~\cite{hu2021lora} fine-tuning in results. Besides, we use a beam search to find optimal fine-tuning hyperparameters, including batch size, learning rate and weight decay of AdamW optimizer~\cite{loshchilov2017decoupled}. Detailed hyperparameters for each dataset can be found in \Cref{appendix:hyper-param}.

\subsection{Remarkable Performance Gain of AsymGQA}
\label{sec:evaluation:asymgqa}

\Cref{tab: 1} presents the accuracy of the grouped LLaMA-2-7B with various average group sizes. We compare AsymGQA to neighbor grouping (``NG''), which evenly groups neighbor key and value layers. Both full fine-tuning results, noted as ``FT'', and LoRA fine-tuning results, noted as ``LoRA'', are included in the table. Another result table of OPT-1.3B can be found in \Cref{tab:opt-1.3b} in~\Cref{appendix:more-results}. We have the following observations.

\begin{itemize}
    \item AsymGQA achieves \textit{consistently higher accuracy than the baseline by a clear margin}, across group sizes and fine-tuning methods. Among these reulsts, the maximum enhancement of accuracy is up to 12.5\%.
    \item This margin (accuracy enhancement) is more obvious on more challenging tasks such as MMLU. For example, full fine-tuned AsymGQA has an average accuracy increase of 6.3\% compared to GQA on MMLU, while on the less challenging SST2, this margin is 2.8\%. 
\end{itemize}

\begin{figure}[H]
\centering
\includegraphics[width=\linewidth]{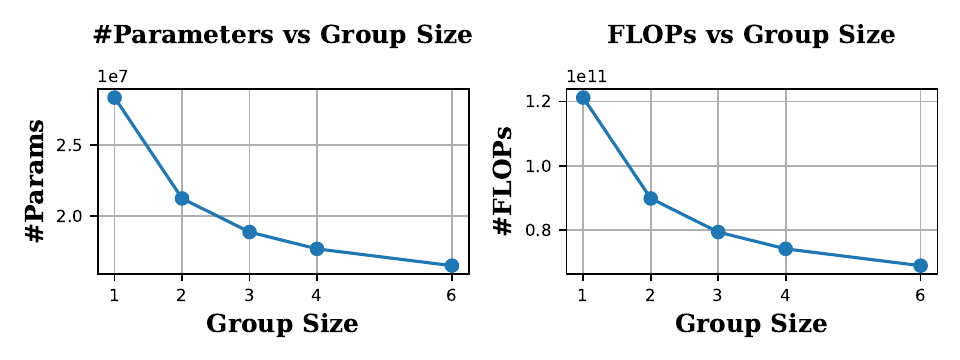}
\vspace{-1em}
\caption{\#Parameters and floating-point operations (FLOPs) vs group size of attention layer. We see a diminishing hardware efficiency return as the group size increases.}
\label{fig: 5}
\end{figure}

We also inspect the trade-off tuned by group size, \textit{i.e.}, trading model quality for hardware efficiency. ~\Cref{fig: 5} illustrates how the number of parameters and inference FLOPs decreases as the group size increases. The hardware efficiency has diminishing returns as we increase the group size. Therefore, for the grouping problem in~\Cref{tab: 1}, a group size of 2 or 3 may be ideal for real-world applications.

\subsection{Ablation Study}
\label{sec:evaluation:ablation-study}

The first ablation study, SG vs NG, shows that even with a uniform group size, our activation-informed grouping (symmetric grouping, noted as ``SG'') still improves the model performance compared to neighbour grouping (NG). The second ablation study shows the performance gain of asymmetric grouping (AG) compared to symmetric grouping.

\begin{figure}[t]
    \centering
    \includegraphics[width=\linewidth]{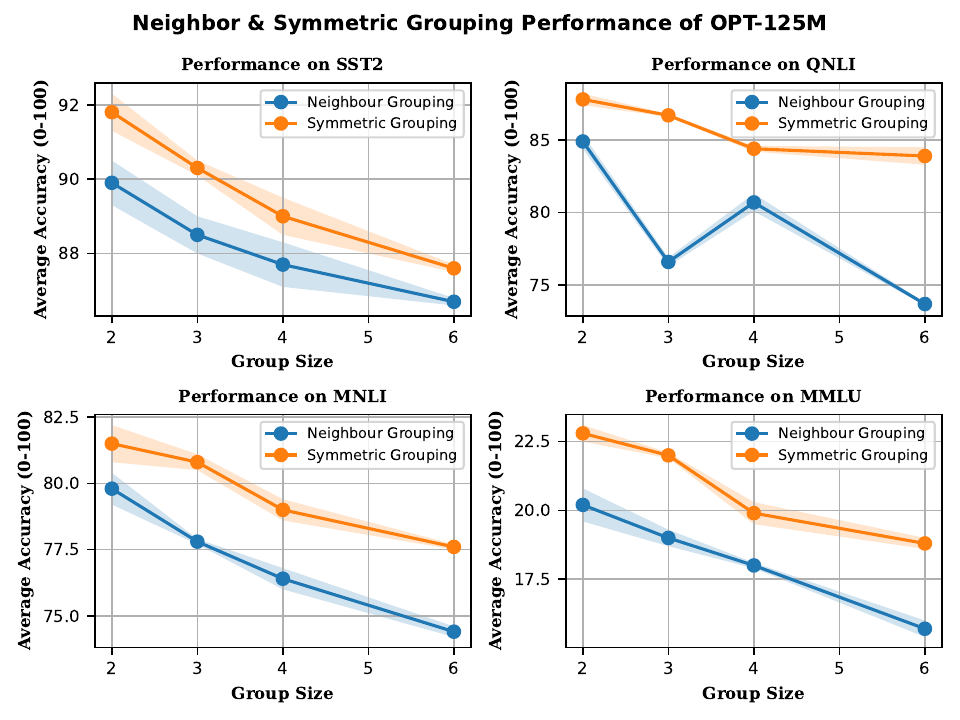}
    \caption{Neighbour grouping vs activation-informed symmetric grouping. Activation-induced similarty between key (value) layers improves model performance even without varied group sizes.}
    \label{fig:sg-vs-ng}
\end{figure}

\paragraph{SG vs NG}~\Cref{fig:sg-vs-ng} compares activation-informed symmetric grouping (SG) to neighbour grouping (NG), indicating that the activation-induced grouping search contributes to performance gain.

\begin{figure}[t]
    \centering
    \includegraphics[width=\linewidth]{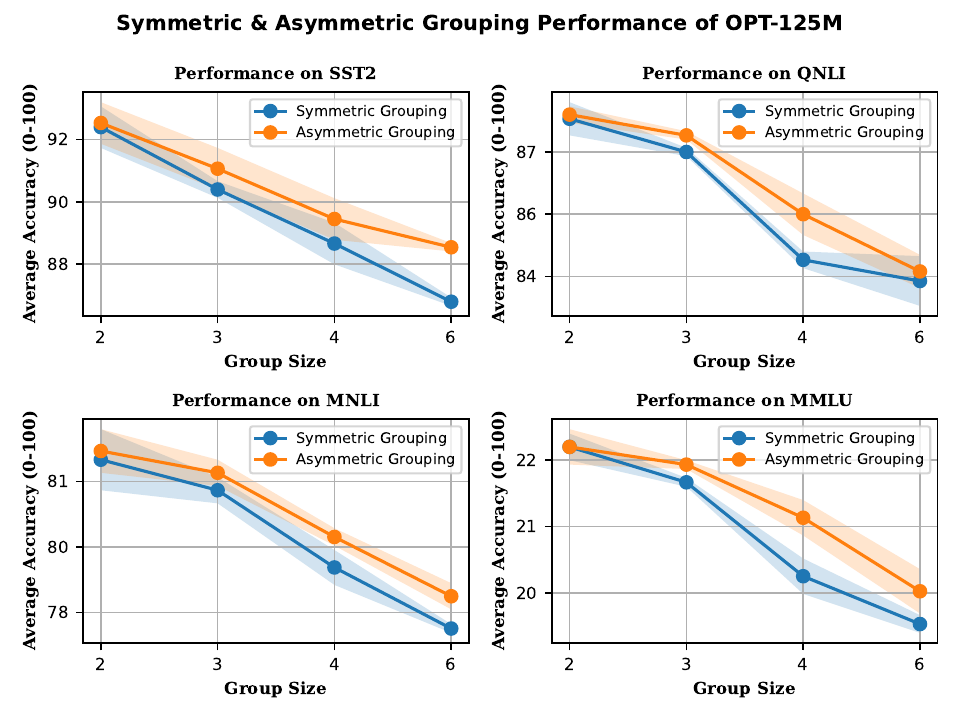}
    \caption{Symmetric vs Asymmetric grouping. Asymmetric further improves model performance by allowing varied group sizes.}
    \label{fig:sg-vs-ag}
\end{figure}

\paragraph{AG vs SG}~\Cref{fig:sg-vs-ag} compares asymmetric grouping (AG) to symmetric grouping (SG), highlighting that varied group size further improves model performance.

\section{Conclusion}
We introduce AsymGQA, an activation-guided asymmetric grouping strategy for transforming a pretrained MHA model into a GQA model. AsymGQA significantly outperforms otherweight-merging baseline, and it  effectively manages the trade-off between model performance and hardware efficiency in GQA.

\nocite{langley00}
\newpage
\bibliography{example_paper}

\begin{thebibliography}{13}
\providecommand{\natexlab}[1]{#1}
\providecommand{\url}[1]{\texttt{#1}}
\expandafter\ifx\csname urlstyle\endcsname\relax
  \providecommand{\doi}[1]{doi: #1}\else
  \providecommand{\doi}{doi: \begingroup \urlstyle{rm}\Url}\fi

\bibitem[Ainslie et~al.(2023)Ainslie, Lee-Thorp, de~Jong, Zemlyanskiy, Lebrón, and Sanghai]{ainslie2023gqa}
Ainslie, J., Lee-Thorp, J., de~Jong, M., Zemlyanskiy, Y., Lebrón, F., and Sanghai, S.
\newblock Gqa: Training generalized multi-query transformer models from multi-head checkpoints, 2023.

\bibitem[Brown et~al.(2020)Brown, Mann, Ryder, Subbiah, Kaplan, Dhariwal, Neelakantan, Shyam, Sastry, Askell, et~al.]{brown2020language}
Brown, T., Mann, B., Ryder, N., Subbiah, M., Kaplan, J.~D., Dhariwal, P., Neelakantan, A., Shyam, P., Sastry, G., Askell, A., et~al.
\newblock Language models are few-shot learners.
\newblock \emph{Advances in neural information processing systems}, 33:\penalty0 1877--1901, 2020.

\bibitem[Devlin et~al.(2019)Devlin, Chang, Lee, and Toutanova]{devlin2019bert}
Devlin, J., Chang, M.-W., Lee, K., and Toutanova, K.
\newblock Bert: Pre-training of deep bidirectional transformers for language understanding, 2019.

\bibitem[Hendrycks et~al.(2020)Hendrycks, Burns, Basart, Zou, Mazeika, Song, and Steinhardt]{DBLP:journals/corr/abs-2009-03300}
Hendrycks, D., Burns, C., Basart, S., Zou, A., Mazeika, M., Song, D., and Steinhardt, J.
\newblock Measuring massive multitask language understanding.
\newblock \emph{CoRR}, abs/2009.03300, 2020.
\newblock URL \url{https://arxiv.org/abs/2009.03300}.

\bibitem[Hu et~al.(2021)Hu, Shen, Wallis, Allen-Zhu, Li, Wang, Wang, and Chen]{hu2021lora}
Hu, E.~J., Shen, Y., Wallis, P., Allen-Zhu, Z., Li, Y., Wang, S., Wang, L., and Chen, W.
\newblock Lora: Low-rank adaptation of large language models, 2021.

\bibitem[Loshchilov \& Hutter(2017)Loshchilov and Hutter]{loshchilov2017decoupled}
Loshchilov, I. and Hutter, F.
\newblock Decoupled weight decay regularization.
\newblock \emph{arXiv preprint arXiv:1711.05101}, 2017.

\bibitem[Socher et~al.(2013)Socher, Perelygin, Wu, Chuang, Manning, Ng, and Potts]{socher2013recursive}
Socher, R., Perelygin, A., Wu, J., Chuang, J., Manning, C.~D., Ng, A.~Y., and Potts, C.
\newblock Recursive deep models for semantic compositionality over a sentiment treebank.
\newblock In \emph{Proceedings of the 2013 conference on empirical methods in natural language processing}, pp.\  1631--1642, 2013.

\bibitem[Touvron et~al.(2023{\natexlab{a}})Touvron, Lavril, Izacard, Martinet, Lachaux, Lacroix, Rozière, Goyal, Hambro, Azhar, Rodriguez, Joulin, Grave, and Lample]{touvron2023llama}
Touvron, H., Lavril, T., Izacard, G., Martinet, X., Lachaux, M.-A., Lacroix, T., Rozière, B., Goyal, N., Hambro, E., Azhar, F., Rodriguez, A., Joulin, A., Grave, E., and Lample, G.
\newblock Llama: Open and efficient foundation language models, 2023{\natexlab{a}}.

\bibitem[Touvron et~al.(2023{\natexlab{b}})Touvron, Martin, Stone, Albert, Almahairi, Babaei, Bashlykov, Batra, Bhargava, Bhosale, Bikel, Blecher, Ferrer, Chen, Cucurull, Esiobu, Fernandes, Fu, Fu, Fuller, Gao, Goswami, Goyal, Hartshorn, Hosseini, Hou, Inan, Kardas, Kerkez, Khabsa, Kloumann, Korenev, Koura, Lachaux, Lavril, Lee, Liskovich, Lu, Mao, Martinet, Mihaylov, Mishra, Molybog, Nie, Poulton, Reizenstein, Rungta, Saladi, Schelten, Silva, Smith, Subramanian, Tan, Tang, Taylor, Williams, Kuan, Xu, Yan, Zarov, Zhang, Fan, Kambadur, Narang, Rodriguez, Stojnic, Edunov, and Scialom]{touvron2023llama2}
Touvron, H., Martin, L., Stone, K., Albert, P., Almahairi, A., Babaei, Y., Bashlykov, N., Batra, S., Bhargava, P., Bhosale, S., Bikel, D., Blecher, L., Ferrer, C.~C., Chen, M., Cucurull, G., Esiobu, D., Fernandes, J., Fu, J., Fu, W., Fuller, B., Gao, C., Goswami, V., Goyal, N., Hartshorn, A., Hosseini, S., Hou, R., Inan, H., Kardas, M., Kerkez, V., Khabsa, M., Kloumann, I., Korenev, A., Koura, P.~S., Lachaux, M.-A., Lavril, T., Lee, J., Liskovich, D., Lu, Y., Mao, Y., Martinet, X., Mihaylov, T., Mishra, P., Molybog, I., Nie, Y., Poulton, A., Reizenstein, J., Rungta, R., Saladi, K., Schelten, A., Silva, R., Smith, E.~M., Subramanian, R., Tan, X.~E., Tang, B., Taylor, R., Williams, A., Kuan, J.~X., Xu, P., Yan, Z., Zarov, I., Zhang, Y., Fan, A., Kambadur, M., Narang, S., Rodriguez, A., Stojnic, R., Edunov, S., and Scialom, T.
\newblock Llama 2: Open foundation and fine-tuned chat models, 2023{\natexlab{b}}.

\bibitem[Vaswani et~al.(2023)Vaswani, Shazeer, Parmar, Uszkoreit, Jones, Gomez, Kaiser, and Polosukhin]{vaswani2023attention}
Vaswani, A., Shazeer, N., Parmar, N., Uszkoreit, J., Jones, L., Gomez, A.~N., Kaiser, L., and Polosukhin, I.
\newblock Attention is all you need, 2023.

\bibitem[Wang et~al.(2018)Wang, Singh, Michael, Hill, Levy, and Bowman]{wang-etal-2018-glue}
Wang, A., Singh, A., Michael, J., Hill, F., Levy, O., and Bowman, S.
\newblock {GLUE}: A multi-task benchmark and analysis platform for natural language understanding.
\newblock In Linzen, T., Chrupa{\l}a, G., and Alishahi, A. (eds.), \emph{Proceedings of the 2018 {EMNLP} Workshop {B}lackbox{NLP}: Analyzing and Interpreting Neural Networks for {NLP}}, pp.\  353--355, Brussels, Belgium, November 2018. Association for Computational Linguistics.
\newblock \doi{10.18653/v1/W18-5446}.
\newblock URL \url{https://aclanthology.org/W18-5446}.

\bibitem[Williams et~al.(2017)Williams, Nangia, and Bowman]{williams2017broad}
Williams, A., Nangia, N., and Bowman, S.~R.
\newblock A broad-coverage challenge corpus for sentence understanding through inference.
\newblock \emph{arXiv preprint arXiv:1704.05426}, 2017.

\bibitem[Zhang et~al.(2022)Zhang, Roller, Goyal, Artetxe, Chen, Chen, Dewan, Diab, Li, Lin, Mihaylov, Ott, Shleifer, Shuster, Simig, Koura, Sridhar, Wang, and Zettlemoyer]{zhang2022opt}
Zhang, S., Roller, S., Goyal, N., Artetxe, M., Chen, M., Chen, S., Dewan, C., Diab, M., Li, X., Lin, X.~V., Mihaylov, T., Ott, M., Shleifer, S., Shuster, K., Simig, D., Koura, P.~S., Sridhar, A., Wang, T., and Zettlemoyer, L.
\newblock Opt: Open pre-trained transformer language models, 2022.

\end{thebibliography}
\bibliographystyle{icml2023}

\newpage
\appendix
\onecolumn
\section{Activation-Informed Grouping Search}
\label{appendix:alg-grouping-search}

\Cref{alg:sym-group} and \Cref{alg:asym-group} describe \textit{activation-informed symmetric grouping} and \textit{activation-informed asymmetric grouping} respectively.

\begin{algorithm}[t]
   \caption{Activation-Informed Symmetric Grouping Search}
   \label{alg:sym-group}
\begin{algorithmic}
    \STATE {\bfseries Input:}
        \vspace{-1em}
        \begin{itemize}\setlength\itemsep{-0.25em}
       \item[] $\text{model}=$ pretrained MHA model
       \item[] $n =$ number of search iterations
       \item[] $m =$ group size
       \item[] $k =$ number of similar attention heads to consider when altering groupings
       \item[] $p_{\text{acc}} =$ probability of accepting a worse-performing group
       \item[] $p_{\text{reset}} =$ probability of resetting the current grouping at each iteration
       \end{itemize}
 
    \FUNCTION{\texttt{SYMMETRIC\_GROUPING} (model, $n$, $m$, $k$, $p_{\text{acc}}$, $p_{\text{reset}}$):}
    \STATE $S \gets \{h | h\text{ is an attention head in model}\}$   
    \STATE $g \gets \{P[i:i+m]|i=0,m,2m,\dots\}, \text{where } P \sim \text{Uniform(permutation(S))}$
    \COMMENT{A random grouping of size $m$.}
    \STATE $\text{best\_acc} \gets 0$
    \COMMENT{Keep track of best accuracy.}
    \STATE $\text{best\_grouping} \gets \text{None}$
    \COMMENT{Keep track of best grouping.}
    \FOR{$i=1$ {\bfseries to} $n-1$}
        \STATE$x \gets X \sim \text{Uniform}(0, 1)$
        \IF{$x < p_{\text{reset}}$}
            \STATE $g \gets \{P[i:i+m]|i=0,m,2m,\dots\}, \text{where } P \sim \text{Uniform(permutation(S))}$
            \COMMENT{Reset $g$ with probability $p_{\text{reset}}$.}
        \ENDIF
        \STATE $a \gets A \sim \text{Uniform}(S)$
        \COMMENT{A randomly chosen attention head.}
        \STATE $H \gets \text{top-}k(S, a)$
        \COMMENT{Top $k$ closest heads in a different group from the group of $a$.}
        \STATE $b \gets B \sim \text{Uniform}(H)$
        \COMMENT{A randomly chosen attention head from $H$.}
        \STATE $g' \gets \text{swap } a \text{ with an element from the group of } b \text{ in } g$
        \COMMENT{Move head $a$ to be grouped with a head similar to $a$, while keeping the equal sizes of all groups.}
        \STATE $y \gets Y \sim \text{Uniform}(0, 1)$
        \IF{$\text{acc}(\text{GQA}(g')) > \text{best\_acc }\textbf{or }  y < p_{\text{acc}}$}
            \STATE $g' \gets g$
            \IF{$\text{acc}(\text{GQA}(g')) > \text{best\_acc}$}
                \STATE $\text{best\_acc} \gets \text{acc}(\text{GQA}(g'))$
                \COMMENT{Update best accuracy.}
                \STATE $\text{best\_grouping} \gets g'$
                \COMMENT{Update best grouping.}
            \ENDIF
        \ENDIF
    \ENDFOR
    \STATE \textbf{return} $\text{best\_acc}, \text{best\_grouping}$
   
   \ENDFUNCTION{}
   
\end{algorithmic}
\end{algorithm}

\begin{algorithm}[H]
\caption{Activation-Informed Asymmetric Grouping Search}
\label{alg:asym-group}
\begin{algorithmic}

    \STATE {\bfseries Input:}
        \vspace{-1em}
        \begin{itemize}\setlength\itemsep{-0.25em}
       \item[] $\text{model}=$ pretrained MHA model
       \item[] $n =$ number of search iterations
       \item[] $m =$ group size
       \item[] $k =$ number of similar attention heads to consider when altering groupings
       \item[] $p_{\text{acc}} =$ probability of accepting a worse-performing group
       \item[] $p_{\text{reset}} =$ probability of resetting the current grouping at each iteration
       \item[] $p_{\text{preserve}} =$ probability of preserving the difference of group sizes when moving a head from one group to another
       \end{itemize}

\FUNCTION{\texttt{SYMMETRIC\_GROUPING} (model, $n$, $m$, $k$, $p_{\text{acc}}$, $p_{\text{reset}}$, $p_{\text{preserve}}$):}
\STATE $S \gets \{h | h\text{ is an attention head in model}\}$
\STATE $g \gets \{P[i:i+m]|i=0,m,2m,\dots\}, \text{where } P \sim \text{Uniform(permutation(S))}$
\COMMENT{\textcolor{gray}{\textit{A random grouping of size $m$.}}}
\STATE $\text{best\_acc} \gets 0$
\COMMENT{\textcolor{gray}{\textit{Best accuracy.}}}
\STATE $\text{best\_grouping} \gets \text{None}$
\COMMENT{\textcolor{gray}{\textit{Keep track of the best-performing grouping during search.}}}
\FOR{$i=0, \dots, n-1$}
\STATE $x \gets X \sim \text{Uniform}(0, 1)$
\COMMENT{\textcolor{gray}{\textit{A random real number from [0,1].}}}
\IF{$x < p_{\text{reset}}$}
\STATE $g \gets \{P[i:i+m]|i=0,m,2m,\dots\}, \text{where } P \sim \text{Uniform(permutation(S))}$
\COMMENT{\textcolor{gray}{Reset $g$ with probability $p_{\text{reset}}$.}}
\ENDIF
\STATE $a \gets A \sim \text{Uniform}(S)$
\COMMENT{\textcolor{gray}{\textit{A randomly chosen attention head.}}}
\STATE $H \gets \text{top-}k(S, a)$
\COMMENT{\textcolor{gray}{\textit{Top $k$ closest heads in a different group from the group of $a$.}}}
\STATE $b \gets B \sim \text{Uniform}(H)$
\COMMENT{\textcolor{gray}{\textit{A randomly chosen attention head from $H$.}}}
\STATE $y \gets Y \sim \text{Uniform}(0, 1)$
\IF{$y < p_{\text{preserve}}$}
\STATE $g' \gets \text{move a head from the group of } b \text{ back to the group of } a \text{ in } g$
\COMMENT{\textcolor{gray}{\textit{Preserve the difference of sizes of two groups.}}}
\ENDIF
\STATE $g' \gets \text{move } a \text{ into the group of } b \text{ in } g$
\STATE $z \gets Z \sim \text{Uniform}(0, 1)$
\IF{$\text{acc}(\text{GQA}(g')) > \text{best\_acc }\textbf{or }  z < p_{\text{acc}}$}
\STATE $g' \gets g$
\COMMENT{\textcolor{gray}{Update $g$ with $g'$.}}
\IF{$\text{acc}(\text{GQA}(g')) > \text{best\_acc}$}
\STATE $\text{best\_acc} \gets \text{acc}(\text{GQA}(g'))$
\COMMENT{Update the best accuracy.}
\STATE $\text{best\_grouping} \gets g'$
\COMMENT{\textcolor{gray}{Update the best-performing grouping.}}
\ENDIF
\ENDIF
\ENDFOR
\STATE \textbf{return} $\text{best\_acc}, \text{best\_grouping}$
\ENDFUNCTION{}
\end{algorithmic}
\end{algorithm}

\section{Weight-Informed Similarity vs Activation-Informed Similarity}
\label{appendix:w-vs-act-sim}

Before diving into Asymmetric search, we perform an experiment to compare the weight-informed layer similarity and the activation-informed similarity. As shown in~\Cref{tab:w-vs-act-similarity}, activation-informed search usually has the best performance.

\begin{table}[H]
\caption{The model performance of brute force search, weight-informed symmetric search, and activation-informed simmetric search. The best is highlighted in bold. Model here is OPT-125M. We find that activation-informed layer similarity usually has the best performance.}
\vspace{1em}
\footnotesize
\centering
\begin{tabular}{@{}cccc@{}}
\toprule
\multicolumn{4}{c}{ACC ($\Delta$) on SST2}\\

{Group Size} & {Brute Force Search} & {Weight-Informed} & {Activation-Informed}\\
\midrule
1 & 92.1±0.1 (0.0) & 92.1±0.1 (0.0)  & 92.1±0.1 (0.0)\\
2 & {91.6±0.3 (-0.5)} & 91.2±0.6 (-0.9) & \textbf{91.8±0.5 (-0.3)}\\
3 & {90.1±0.5 (-2.0)} & 90.0±0.1 (-2.1) & \textbf{90.3±0.2 (-1.8)}\\
4 & 88.9±0.1 (-3.2) & \textbf{89.1±0.4 (-3.0)} & {89.0±0.5 (-3.1)}\\
6 & 87.5±0.1 (-4.6) & \textbf{87.8±0.4 (-4.3)} & {87.6±0.1 (-4.5)}\\
\midrule
\multicolumn{4}{c}{ACC ($\Delta$) on QNLI}\\

{Group Size} & {Brute Force Search} & {Weight-Informed}  & {Activation-Informed}\\
\midrule
1 & 88.1±0.4 (0.0) & 88.1±0.3 (0.0)  & 88.1±0.3 (0.0)\\
2 & 87.4±0.6 (-0.7) & {87.7±0.1 (-0.4)}  & \textbf{87.8±0.4 (-0.3)}\\
3 & {86.6±0.4 (-1.5)} & 85.6±0.1 (-2.5)  & \textbf{86.7±0.1 (-1.4)}\\
4 & 84.0±0.1 (-4.1) & \textbf{84.8±0.4 (-3.3)}  & 84.4±0.2 (-3.7)\\
6 & 82.5±0.2 (-5.6) & 83.4±0.1 (-4.7)  & \textbf{83.9±0.6 (-4.2)}\\

\midrule
\multicolumn{4}{c}{ACC ($\Delta$) on MNLI}\\

{Group Size} & {Brute Force Search} & {Weight-Informed}& {Activation-Informed}\\
\midrule
1 & 82.1±0.2 (0.0) & 82.1±0.1 (0.0)  & 82.1±0.1 (0.0)\\
2 & \textbf{{81.4±0.2 (-0.7)}} & \textbf{{81.4±0.1 (-0.7)}} & \textbf{81.5±0.7 (-0.6)}\\
3 & 80.3±0.1 (-1.8) & 80.4±0.3 (-1.7) & \textbf{80.8±0.3 (-1.3)}\\
4 & 78.5±0.5 (-3.6) & 78.4±0.2 (-3.7)  & \textbf{79.0±0.4 (-3.1)}\\
6 & 77.6±0.6 (-4.5) & \textbf{77.8±0.5 (-4.4)} & 77.6±0.1 (-4.5)\\

\midrule
\multicolumn{4}{c}{ACC ($\Delta$) on MMLU}\\

{Group Size} & {Brute Force Search} & {Weight-Informed}  & {Activation-Informed}\\
\midrule
1 & 23.1±0.3 (0.0) & 23.1±0.1 (0.0)  & 23.1±0.1 (0.0)\\
2 & 22.5±0.1 (-0.6) & \textbf{22.8±0.6 (-0.3)}  & \textbf{22.8±0.3 (-0.3)}\\
3 & 21.6±0.4 (-1.5) & {21.7±0.4 (-1.4)}  & \textbf{22.0±0.1 (-1.1)}\\
4 & 19.7±0.5 (-3.4) & \textbf{20.0±0.5 (-3.1)}  & {19.9±0.4 (-3.2)}\\
6 & {18.5±0.4 (-4.6)} & 18.2±0.6 (-4.9)  & \textbf{18.8±0.2 (-4.3)}\\

\bottomrule
\end{tabular}
\label{tab:w-vs-act-similarity}
\end{table}

\section{Hyperparameters}
\label{appendix:hyper-param}

\begin{table}[H]
\caption{Details of the selected hyper-parameters, including batch size, learning rate $\eta$ and
weight decay $w_d$ for each set of experiments with the same dataset and fine-tuning method.}
\vspace{1em}
\centering
\begin{tabular}{@{}ccccc|ccccc@{}}
\toprule

\multicolumn{1}{c}{\textbf{Dataset}} & \multicolumn{1}{c}{\textbf{LoRA}} & \multicolumn{1}{c}{\textbf{Batch Size}} &  \multicolumn{1}{c}{\textbf{$\eta$}} & \multicolumn{1}{c}{\textbf{$w_d$}} & \multicolumn{1}{c}{\textbf{Dataset}} & \multicolumn{1}{c}{\textbf{LoRA}} & \multicolumn{1}{c}{\textbf{Batch Size}} &  \multicolumn{1}{c}{\textbf{$\eta$}} & \multicolumn{1}{c}{\textbf{$w_d$}}\\
\midrule

\multirow{1}{*}{SST2} & \parbox{1cm}{\centering No \\ Yes} & \parbox{2cm}{\centering 256 \\ 128} & \parbox{0.8cm}{\centering 1e-5 \\ 2e-5} & \parbox{0.8cm}{\centering 0.01 \\ 0.01} & \multirow{1}{*}{QNLI} & \parbox{1cm}{\centering No \\ Yes} & \parbox{2cm}{\centering 128 \\ 128} & \parbox{1cm}{\centering 2e-5 \\ 2e-5} & \parbox{1cm}{\centering 0.1 \\ 0.1}\\
\midrule

\multirow{1}{*}{MNLI} & \parbox{1cm}{\centering No \\ Yes} & \parbox{2cm}{\centering 128 \\ 128} & \parbox{0.8cm}{\centering 2e-5 \\ 1e-5} & \parbox{0.8cm}{\centering 0.01 \\ 0.01} & \multirow{1}{*}{Alpaca} & \parbox{1cm}{\centering No \\ Yes} & \parbox{2cm}{\centering 128 \\ 128} & \parbox{1cm}{\centering 2e-5 \\ 2e-5} & \parbox{1cm}{\centering 0.01 \\ 0.01}\\

\bottomrule
\label{tab: A.2}
\end{tabular}
\vspace{-5mm}
\end{table}

\section{More Experiment Results}
\label{appendix:more-results}

We also run the experiment to compare neighbour grouping and AsymGQA on OPT-1.3B. As shown in~\Cref{tab:opt-1.3b}, AsymGQA still outperforms GQA by a clear margin.

\begin{table}[H]
\caption{The performance of OPT-1.3B with experimentally best GQA architecture (i.e. asymmetric grouping + random-search-based approach + vector-wise cosine similarity on activations) compared with neighbour grouping. Column \texttt{NG} means neighbour grouping. Columns noted with \texttt{LoRA} are results of models fine-tuned by LoRA while other columns are from models fully fine-tuned. Grouping methods with the highest accuracies are highlighted in bold.}
\footnotesize
\centering
\begin{tabular}{@{}ccccc@{}}
\toprule
\multicolumn{5}{c}{ACC ($\Delta$) on SST2}\\

\textbf{Group Size} & GQA (FT) & AsymGQA (FT) & GQA (LoRA) & AsymGQA (LoRA)\\
\midrule
1 & 93.0±0.5 (0.0) & 93.0±0.5 (0.0) & 92.8±0.6 (0.0) & 92.8±0.6 (0.0)\\
2 & 91.3±0.5 (-1.7) & \textbf{92.8±0.7 (-0.2)} & 89.6±0.1 (-3.1) & \textbf{91.9±0.5 (-0.8)}\\
3 & 89.2±0.5 (-3.8) & \textbf{91.9±0.3 (-1.1)} & 88.1±0.2 (-4.7) & \textbf{91.2±0.2 (-1.5)}\\
4 & 88.5±0.6 (-4.5) & \textbf{90.6±0.2 (-2.4)} & 86.6±0.4 (-5.2) & \textbf{90.2±0.1 (-2.5)}\\
6 & 87.7±0.2 (-5.3) & \textbf{90.5±0.5 (-2.5)} & 84.6±0.1 (-8.2) & \textbf{89.1±0.6 (-3.7)}\\
\midrule
\multicolumn{5}{c}{ACC ($\Delta$) on QNLI}\\

\textbf{Group Size} & GQA (FT) & AsymGQA (FT) & GQA (LoRA) & AsymGQA (LoRA)\\
\midrule
1 & 89.1±0.3 (0.0) & 89.1±0.3 (0.0) & 89.0±0.7 (0.0) & 89.0±0.7 (0.0)\\
2 & 84.7±0.2 (-4.4) & \textbf{88.7±0.1 (-0.4)} & 84.3±0.1 (-4.7) & \textbf{88.2±0.4 (-0.8)}\\
3 & 84.2±0.4 (-4.9) & \textbf{87.3±0.1 (-1.8)} & 82.8±0.6 (-6.2) & \textbf{87.2±0.4 (-1.8)}\\
4 & 79.1±0.5 (-10.0) & \textbf{85.7±0.5 (-3.4)} & 80.3±0.4 (-8.7) & \textbf{85.1±0.6 (-3.9)}\\
6 & 78.0±0.1 (-11.1) & \textbf{85.0±0.4 (-4.1)} & 77.9±0.6 (-11.1) & \textbf{84.2±0.2 (-4.8)}\\

\midrule
\multicolumn{5}{c}{ACC ($\Delta$) on MNLI}\\

\textbf{Group Size} & GQA (FT) & AsymGQA (FT) & GQA (LoRA) & AsymGQA (LoRA)\\
\midrule
1 & 84.2±0.6 (0.0) & 84.2±0.6 (0.0) & 83.8±0.5 (0.0) & 83.8±0.5 (0.0)\\
2 & 81.3±0.2 (-2.9) & \textbf{83.7±0.3 (-0.5)} & 80.1±0.5 (-3.7) & \textbf{81.8±0.3 (-1.0)}\\
3 & 79.4±0.5 (-4.8) & \textbf{82.8±0.2 (-1.4)} & 78.9±0.3 (-4.9) & \textbf{82.5±0.6 (-1.3)}\\
4 & 79.0±0.2 (-5.2) & \textbf{81.2±0.4 (-3.0)} & 77.9±0.5 (-5.9) & \textbf{81.1±0.2 (-2.7)}\\
6 & 76.7±0.3 (-7.5) & \textbf{80.0±0.1 (-4.2)} & 75.3±0.5 (-8.5) & \textbf{79.2±0.6 (-4.6)}\\

\midrule
\multicolumn{5}{c}{ACC ($\Delta$) on MMLU}\\

\textbf{Group Size} & GQA (FT) & AsymGQA (FT) & GQA (LoRA) & AsymGQA (LoRA)\\
\midrule
1 & 22.9±0.3 (0.0) & 22.9±0.3 (0.0) & 23.0±0.2 (0.0) & 23.0±0.2 (0.0)\\
2 & 21.6±0.5 (-1.3) & \textbf{23.2±0.1 (0.3)} & 20.3±0.6 (-2.7) & \textbf{22.6±0.7 (-0.4)}\\
3 & 19.7±0.1 (-3.2) & \textbf{22.3±0.7 (-0.6)} & 19.5±0.2 (-3.5) & \textbf{21.8±0.1 (-1.2)}\\
4 & 17.5±0.4 (-5.4) & \textbf{20.7±0.5 (-2.2)} & 16.5±0.5 (-6.5) & \textbf{20.9±0.6 (-2.1)}\\
6 & 16.5±0.6 (-6.4) & \textbf{19.2±0.5 (-3.6)} & 15.2±0.4 (-7.8) & \textbf{19.4±0.7 (-3.6)}\\

\bottomrule
\end{tabular}
\label{tab:opt-1.3b}
\end{table}

\end{document}